\documentclass[letterpaper, 10 pt, conference]{ieeeconf}  

\IEEEoverridecommandlockouts                              

\usepackage{epsfig} 
\usepackage{times} 
\usepackage{amsmath} 
\usepackage{amssymb}  
\usepackage{color}
\usepackage{xcolor}
\usepackage{preambles}
\usepackage{amsmath}

\usepackage{amsthm}
\usepackage{graphicx}
\usepackage{url} 
\usepackage[format=plain,labelsep=period]{caption}
\usepackage{wrapfig}
\usepackage[list=on,listformat=simple]{subcaption}
\usepackage{hyperref}
\usepackage{url}
\usepackage{makecell}
\usepackage{tabularray}
\usepackage{pbox}
\usepackage{caption}
\usepackage{float}
\floatstyle{plaintop}
\restylefloat{table}
\usepackage{subcaption}
\usepackage{mathtools}
\hypersetup{colorlinks = true, citecolor=black}
    
\usepackage{flushend}

\title{\LARGE \bf
Real-Time Safe Bipedal Robot Navigation using Linear Discrete Control Barrier Functions
}
\author{Chengyang Peng$^{1 \dagger}$, Victor Paredes$^{1 \dagger}$,  Guillermo A. Castillo$^{2}$ and Ayonga Hereid$^{1}$
\thanks{This work was supported in part by the National Science Foundation under grant FRR-21441568. }%
\thanks{$^{1}$Mechanical and Aerospace Engineering, The Ohio State University, Columbus, OH, USA. {\tt\footnotesize (peng.947, paredescauna.1, hereid.1)@osu.edu.}}%
\thanks{$^{2}$Electrical and Computer Engineering, The Ohio State University, Columbus, OH, USA;  {\tt\footnotesize castillomartinez.2@osu.edu}}
\thanks{$\dagger$ These authors contributed equally.}
}

\usepackage[capitalise]{cleveref}

\newtheorem*{proposition}{Proposition}
\usepackage[noadjust]{cite}

\usepackage{algorithm2e,algorithmic}

\begin{document}

\maketitle

\begin{abstract}


Safe navigation in real-time is an essential task for humanoid robots in real-world deployment. Since humanoid robots are inherently underactuated thanks to unilateral ground contacts, a path is considered safe if it is obstacle-free and respects the robot's physical limitations and underlying dynamics. Existing approaches often decouple path planning from gait control due to the significant computational challenge caused by the full-order robot dynamics. In this work, we develop a unified, safe path and gait planning framework that can be evaluated online in real-time, allowing the robot to navigate clustered environments while sustaining stable locomotion. Our approach uses the popular Linear Inverted Pendulum (LIP) model as a template model to represent walking dynamics. It incorporates heading angles in the model to evaluate kinematic constraints essential for physically feasible gaits properly. In addition, we leverage discrete control barrier functions (DCBF) for obstacle avoidance, ensuring that the subsequent foot placement provides a safe navigation path within clustered environments. To guarantee real-time computation, we use a novel approximation of the DCBF to produce linear DCBF (LDCBF) constraints. We validate the proposed approach in simulation using a Digit robot in randomly generated environments. The results demonstrate that our approach can generate safe gaits for a non-trivial humanoid robot to navigate environments with randomly generated obstacles in real-time.

\end{abstract}

\section{Introduction}

Humanoid robots exhibit great dexterity and agility to perform different locomotion activities~\cite{Wight2008Foot, hildebrandt2017real, liu2023realtime}. They are expected to be deployed to real-life environments such as warehouses and assembly lines. Such clustered environments pose the challenge of navigating around obstacles while maintaining stable walking in real-time~\cite{wermelinger2016navigation, kuindersma2016optimization}. 
Yet humanoid robots are inherently underactuated due to their unilateral ground contacts; thus, a strong coupling exists between path planning and gait control. Their high-dimensional, nonlinear, and hybrid dynamics further complicate real-time motion planning. 
For fully actuated legged robots, one can decouple the path planning problem from motion control by finding a collision-free path without accounting for the robot dynamics and motion control~\cite{sleumer1999exact,gasparetto2015path}. Then, a feedback controller aware of the robot dynamics can be developed to track the provided path. However, under-actuated humanoids would fall if the planned path did not account for the robot dynamics. This coupling of planning and dynamics usually requires solving specific gait optimization problems based on the robot's full-order ~\cite{dai2014whole, diehl2006fast,mombaur2009using} or reduced-order model~\cite{wieber2006trajectory, garcia2021mpc, liu2023realtime}.
However, using the full-order model for long-horizon path planning requires significant computation time, making them non-amenable for real-time online planning.

\begin{figure}
\centering
\vspace{2mm}
    \includegraphics[trim={2cm 0cm 2cm 0cm},clip,width=1\columnwidth]{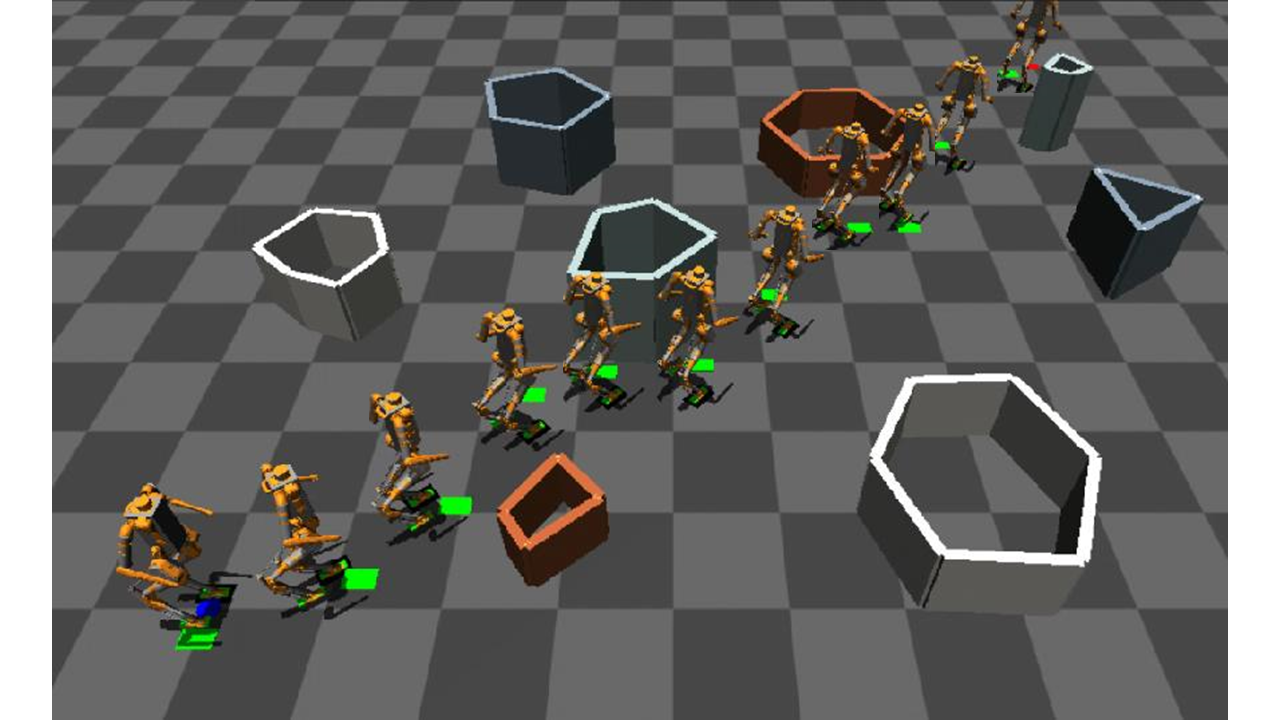}
\caption{\small{Safe navigation planning is tested in MuJoCo with the Digit robot.}
} 
\label{fig:simu_nav_1}
\vspace{-3mm}
\end{figure}

To mitigate the computational challenge, reduced-order template models are often used to approximate the walking dynamics of the robot and plan gaits with reduced computational burden~\cite{teng2021toward, griffin2016model, peng2024unified}. 
A famous example of such template models is the linear inverted pendulum (LIP) model~\cite{kajita20013d}.
In particular, the LIP model represents the robot dynamics using the center of mass (CoM) position and velocity and allows the control of the CoM velocity at the end of the next step.
The LIP model provides a lower-dimensional representation of the robot dynamics that can be used to plan a desired foot placement that renders stable walking gaits. 

In this work, we introduce a modified 3D-LIP model with heading angles to use the LIP model for both path and gait planning and enforce necessary kinematic constraints. Including heading angles and turning rates allows for expressing kinematic constraints in the local coordinate frame of the robot, appropriately addressing the different motion constraints in sagittal and frontal planes, as well as left foot stance and right foot stance, of walking robots. A model predictive control based on the discrete-time step-to-step dynamics of this 3D-LIP model, denoted as LIP-MPC, is proposed to unify path and gait planning. 
In particular, our MPC formulation uses discrete control barrier functions (DCBFs) for safe obstacle avoidance.
Control barrier functions have been successfully applied to controlling legged robots and are now widely used to ensure safety in path planning~\cite{ames2016control, ames2019control, manjunath2021safe, peng2023safe}. DCBF, particularly, is well-suited for the discrete-time step-to-step dynamic model~\cite{agrawal2017discrete, zeng2021safety}. However, many barrier functions that describe obstacles are nonlinear, leading to nonlinear constraints when foot placement is treated as the decision variable. 
Hence, despite the 3D-LIP model being linear, the nonlinearity in both kinematic and path constraints hinders the real-time computation of the MPC problem.


To mitigate computational efficiency, this paper proposes two key ingredients: pre-computing heading angles and approximated linear DCBFs.
Preprocessing the turning rate for each prediction step allows us to linearize the kinematic constraints within the MPC. Additionally, we introduce a novel linear discrete-time Control Barrier Function (LDCBF) to establish linear, feasible obstacle avoidance constraints for convex obstacles. These adjustments transform the optimization problem into a linearly constrained quadratic programming (QP) problem, significantly reducing computational demands and enabling real-time, safety-critical navigation for bipedal robots.

The rest of the paper is organized as follows: Section~\ref{sec:safe-mpc} introduces the 3D-LIP model with heading angles and presents the formulation of the LIP-MPC with linear kinematic constraints for feasible gait planning.  
Section~\ref{sec:dcbf} details the design of the obstacle avoidance constraints, introducing the novel LDCBF expression. Section~\ref{sec:results} shows the application of the proposed LIP-MPC in simulation, showcasing the real-time safe navigation of the bipedal robot Digit. We presented the results of two different turning rate preprocessing strategies: global goal-oriented and subgoal-oriented. Finally, Section~\ref{sec:conclusion} concludes the contributions, limitations, and future work.

\section{Real-Time Gait Planing with LIP-MPC}
\label{sec:safe-mpc}


The full-order dynamics of bipedal robots are high-dimensional, nonlinear, and hybrid, which poses significant computational challenges for planning and control. In this section, we introduce a three-dimensional linear inverted pendulum (3D-LIP) model with heading angles, and formulate a model predictive control formulation that uses the step-to-step discrete dynamics for step planning. We propose several novel measures to ensure kinematics and safety constraints can be represented as linear constraints, thereby allowing real-time gait generation in crowded environments.


\subsection{3D-LIP Model with Heading Angle}


The LIP model assumes the robot keeps a constant center of mass (CoM) height $H$ during the walking step. The motion of the robot along the $x$ direction can be expressed as: 
\begin{align}
\dot{v}_{x} = \frac{g}{H}(p_x-f_x),
\label{eq:lip-continuous}
\end{align}
where $(p_x, v_x)$ are the CoM position and velocity in $x$-axis, $f_x$ is the stance foot position, and $g$ is the gravitational acceleration. If we assume that each walking step takes a fixed time $T$, the closed-form solution of the step-to-step discrete dynamics can be written as:
\begin{align}
\label{eq:x_dir_dynamics}
    \begin{bmatrix} p_{x_{k+1}} \\ v_{x_{k+1}} \end{bmatrix} =  \mathbf{A_d}
    \begin{bmatrix} p_{x_k} \\ v_{x_k} \end{bmatrix}
    + \mathbf{B_d}
    f_{x_k},
\end{align}
with 
\begin{equation}
\label{eq:matrix_small}
\begin{aligned}
    \mathbf{A_d} &\coloneqq \left[ {\begin{array}{cc}
    \cosh(\beta T) & \frac{\sinh(\beta T)}{\beta}\\
   \beta\sinh(\beta T)  & \cosh(\beta T)
    \end{array} } \right],\\
    \mathbf{B_d} &\coloneqq \begin{bmatrix} 1-\cosh(\beta T) \\-\beta\sinh(\beta T) \end{bmatrix},
\end{aligned}
\end{equation}
where $\beta = \sqrt{g/H}$, and $p_{x_k}$ and $v_{x_k}$ represent the CoM position and velocity at the beginning of $k$-th step.

The motion in the y-axis direction has the same expression as the x-axis. In this work, we will also consider the heading angle $\theta$ and its turning rate $\omega$, which can be used to determine the orientation of the local robot coordinates. This allows kinematics constraints (e.g., body velocity, leg reachability, etc.) to be expressed properly in our gait planning problem. 
By defining the state $\mathbf{x} \coloneqq [p_x, v_x, p_y, v_y, \theta]^T\in\mathcal{X}\subset\mathbb{R}^5$ and the control variable $\mathbf{u} \coloneqq [f_x, f_y, \omega]^T\in\mathcal{U}\subset\mathbb{R}^3$, the step-to-step dynamics of the 3D-LIP model can be written as:
\begin{align}
\label{eq:system_dynamics}
    \mathbf{x}_{k+1} =  \mathbf{A_L}
    \mathbf{x}_k
    + \mathbf{B_L}\mathbf{u}_k,
\end{align}
with: 
\begin{equation}
\label{eq:system_matrix}
\begin{aligned}
    \mathbf{A_L} &\coloneqq \begin{bmatrix}
    \mathbf{A_d} & \mathbf{0} & \mathbf{0}\\
    \mathbf{0}& \mathbf{A_d} & \mathbf{0}\\
    \mathbf{0} &  \mathbf{0} & 1
    \end{bmatrix}, \quad
    \mathbf{B_L} &\coloneqq \begin{bmatrix} 
    \mathbf{B_d} & \mathbf{0} & \mathbf{0}\\
    \mathbf{0} & \mathbf{B_d}  & \mathbf{0}\\
    0 & 0 & T
    \end{bmatrix}.
\end{aligned}
\end{equation}

\begin{figure}
\centering
\vspace{2mm}
    \includegraphics[trim={0cm 0cm 0cm 0cm},clip,width=1.0\columnwidth]{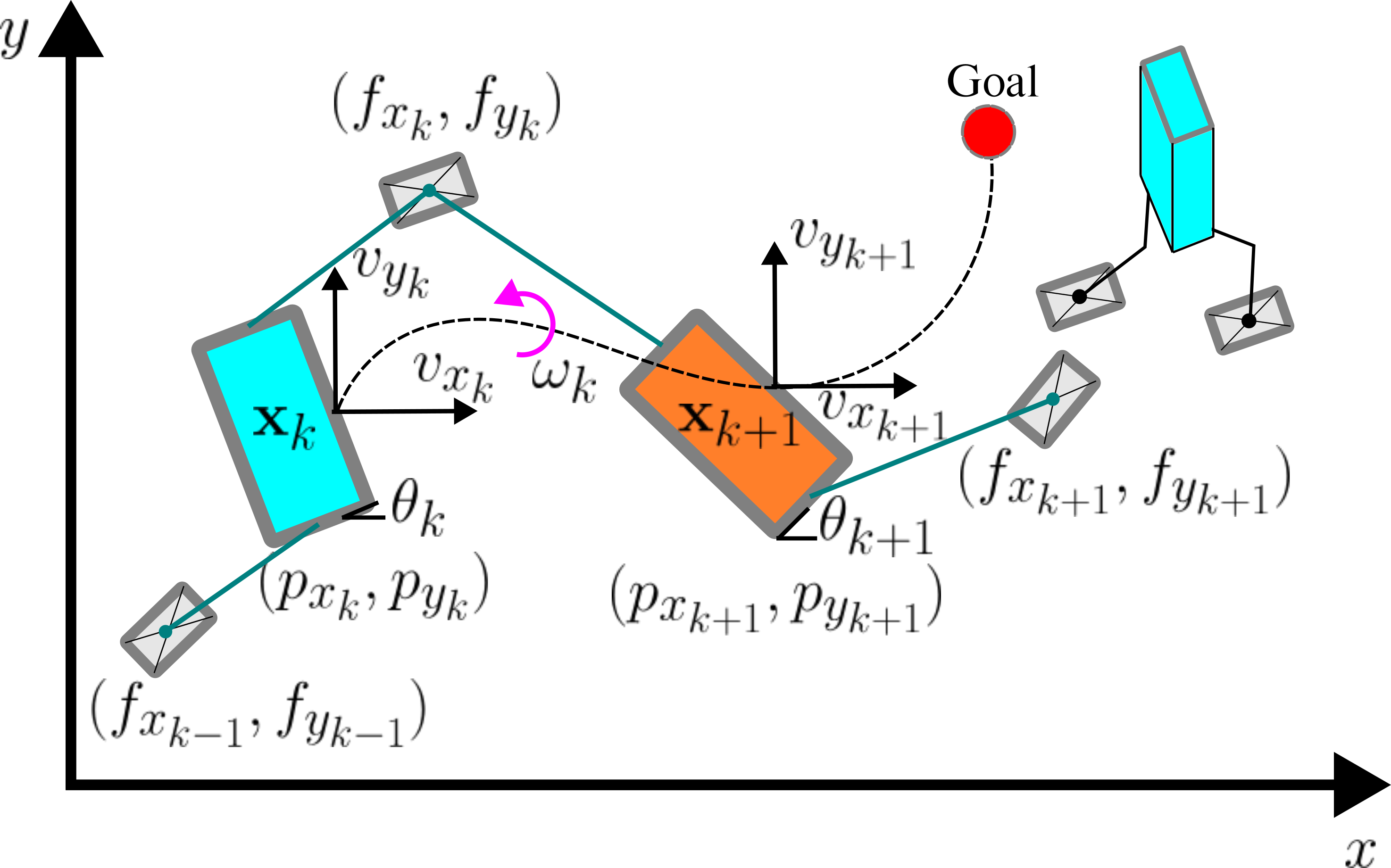}
\caption{Representation of state transition of the LIP model in the global frame. The state  $\mathbf{x}_{k}$ in step $k$ (shown in cyan) evolves to the state $\mathbf{x}_{k+1}$ (shown in orange) when stepping at point $(f_{x_k}, f_{y_k})$ and with a turning rate of $\omega_k$.}
\label{fig:LIP_new}
\vspace{-3mm}
\end{figure}

We show in \figref{fig:LIP_new} the projection of the LIP states and controls in the $x-y$ plane during multiple steps.

\subsection{Safe Gait Planning with Model Predictive Control}


Our work uses the step-to-step 3D-LIP dynamics in \eqref{eq:system_dynamics} to formulate a Model Predictive Control (MPC) problem to compute optimal stepping positions for stable locomotion and safe navigation. To be able to respond to the instantaneous states of the robot in real time, we compute the next discrete states (i.e., the LIP states at the beginning of a step) using the closed form solution of \eqref{eq:lip-continuous}. Let us denote it as $\mathbf{x}_0$, which will serve as the starting point in the MPC formulation. 
Thus, the LIP-MPC can be stated as:
\begin{align}
\label{eq:MPC_prob}
    J^* =& \min_{\mathbf{u}_{0:N-1}} \hspace{1em}  \sum_{k=1}^{N} q(\mathbf{x}_{k}) \\
    \st &\mathbf{x}_{k} \in \mathcal{X}, \text{ } k \in [1, N]\nonumber\\
    &\mathbf{u}_{k} \in \mathcal{U},\text{ } k \in [0, N-1]\nonumber\\
    &\mathbf{x}_{k+1} =  \mathbf{A_L}
    \mathbf{x}_{k}
    + \mathbf{B_L}\mathbf{u}_{k}, \text{ } k \in [0, N-1]\nonumber\\
    &\mathbf{c}_l \leq \mathbf{c}(\mathbf{x}_{k}, \mathbf{u}_{k}) \leq \mathbf{c}_u, \text{ } k \in [0, N-1]\nonumber
\end{align}
where, $\mathcal{X}$ is the set of allowed states and $\mathcal{U}$ is the set of admissible controls. 
$q(\mathbf{x}_k)$ is the cost function, defined to evaluate the distance of a sequence of predicted states from the goal position $(g_{x}, g_y)$. Minimizing this cost implies that the robot moves towards the goal position. In particular, we consider quadratic cost function, defined as:
\begin{align}
\label{eq:cost_func_qp_form}
    q(\mathbf{x}_{k})=&  \left(p_{x_k}-g_{x}\right)^2+\left(p_{y_k}-g_{y}\right)^2 \quad \forall k\in[1,N].
\end{align}
The kinematics and path constraints will be captured in $\mathbf{c}(\mathbf{x}_{k}, \mathbf{u}_{k})$. These constraints are often nonlinear as they must be evaluated in the local coordinate frame, which hinders the real-time computation of \eqref{eq:MPC_prob}. In the following discussion, we introduce several novel measures to linearize these constraints, thereby ensuring the optimization problem in \eqref{eq:MPC_prob} can be solved in real-time.

\begin{figure}
\centering
\vspace{2mm}
    \includegraphics[trim={0cm 0cm 0cm 0cm},clip,width=1\columnwidth]{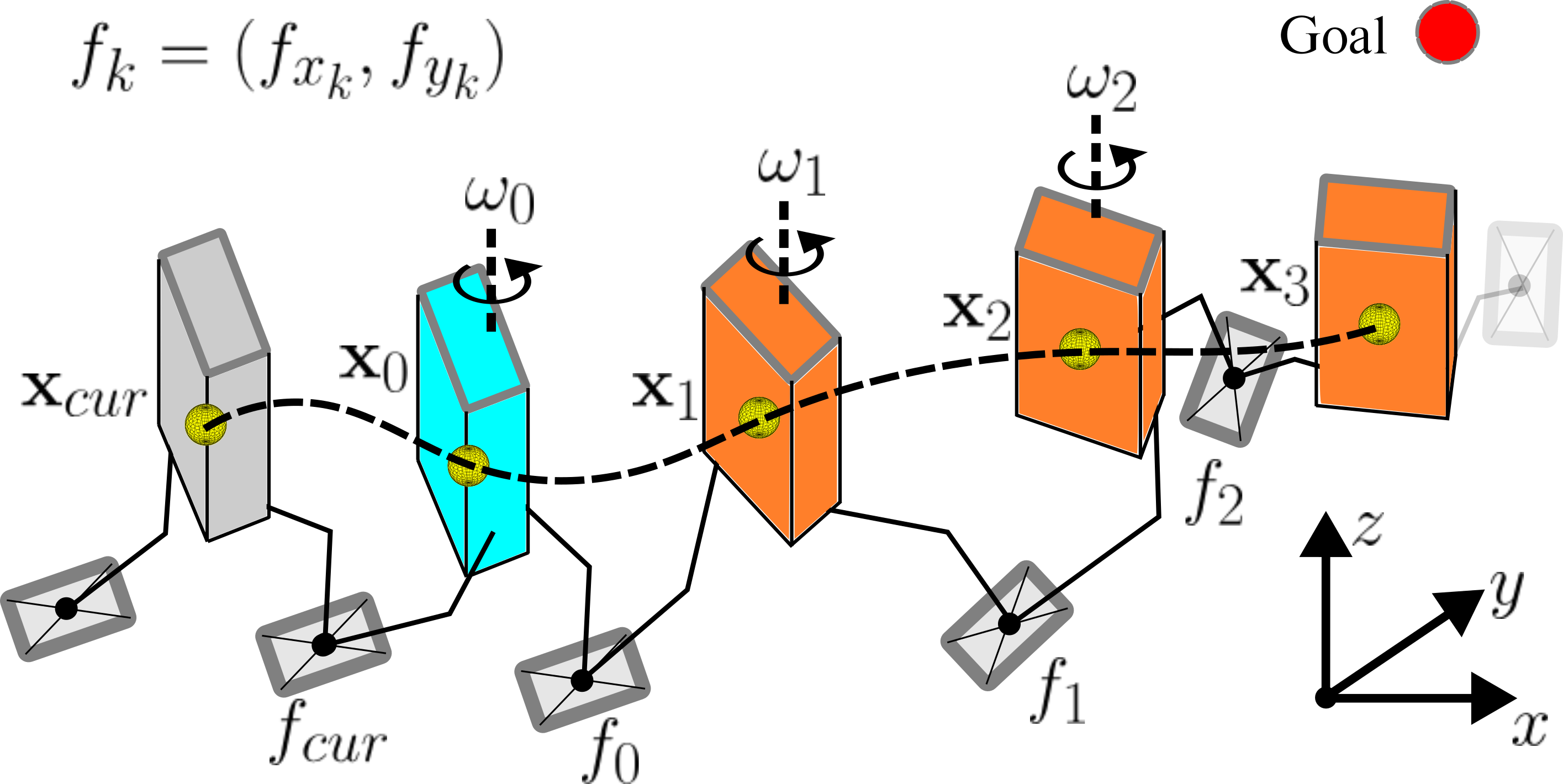}
\caption{An illustration of the LIP-MPC formulation when $N=3$. The planner first estimates the state at the end of the current step, $\mathbf{x}_0$, given the current instantaneous state, $\mathbf{x}_{cur}$, and the stance foot position contained in  $\mathbf{u}_{cur}$. The turning rates in the subsequent steps are pre-computed, where the stepping positions $f_k$ will be optimized by the LIP-MPC.}
\label{fig:3_step_pre}
\vspace{-3mm}
\end{figure}

\subsection{Heading Angle Preprocessing}
The nonlinearity in $\mathbf{c}(\mathbf{x}_k, \mathbf{u}_k)$ is often due to the inclusion of the heading angle, $\theta_k$. 
To linearize these constraints, we propose to pre-compute the turning rates $\bar{\omega}_k, \forall k \in [0,1,...,N-1]$ to keep them fixed in the MPC calculation. A straightforward and simple way is to calculate the target heading angle as the direction from the current position towards the goal position. The required turning rate $\bar{\omega}_{k}$ is calculated by smoothly turning from the current heading to the target heading angle in $N$ steps. To avoid sharp turns, we also impose a limit on the turning rate $|\bar{\omega}_{k}| \leq  0.156\pi \,\mathrm{rad}/s$. \figref{fig:3_step_pre} shows the case when MPC prediction step $N = 3$, and the fixed turning rates for all steps. 

\subsection{Linearized Safety Constraints}
In addition to the obstacle avoidance constraints, which will be presented in the next section, we enforce multiple kinematics constraints to ensure that the optimized stepping positions are physically feasible on the robot hardware. With pre-computed heading angles, these constraints will become linear inequality constraints, as discussed below.




\newsec{Walking Velocities} 
Since the 3D-LIP states in \eqref{eq:system_dynamics} are expressed in the world coordinates, to properly limit the walking velocities, we must compute the body velocities in the robot's local coordinates. In particular, the walking velocity constraint can be expressed as:
\begin{align}
\label{eq:rigt_vel_cons_rigt}
    \begin{bmatrix} v_{x_\mathrm{min}}\\ v_{y_\mathrm{min}} \end{bmatrix} \leq  \begin{bmatrix} \cos\theta_k & \sin\theta_k \\ -\sin\theta_k &  \cos\theta_k\end{bmatrix}
    \begin{bmatrix} v_{x_{k+1}} \\ s_v v_{y_{k+1} }\end{bmatrix}
    \leq \begin{bmatrix} v_{x_\mathrm{max}}\\ v_{y_\mathrm{max}} \end{bmatrix}
\end{align}
$\forall k \in [0, N-1]$,
where, $v_{x_{\mathrm{min}}}, v_{x_{\mathrm{max}}}, v_{y_{\mathrm{min}}}, v_{y_{\mathrm{max}}}$ are the lower bounds and upper bounds of the robot longitudinal and lateral velocities, respectively. $s_v$ is a sign function that depends on which foot is the stance. If the right foot is the stance, then $s_v = 1$; otherwise, $s_v = -1$. 
This allows lateral velocity at the end of each step to be properly limited to ensure that the foot lands on the opposite side, thereby preventing potential leg crossing and collisions.

\newsec{Leg Reachability.} The swing foot reachability constraint is used to prevent over-extension of the swing leg. Our previous work~\cite{peng2024unified} limits the Euclidean distance between the robot's center of mass (CoM) and the subsequent stepping position. Despite being straightforward, it introduces nonlinearity in the optimization. We reformulate the constraint to decompose the Euclidean distance into longitudinal and lateral components based on the local coordinate frame. This allows a linear expression of the reachability constraint, given as:
\begin{align}
\label{eq:leg_len_cons}
    &\begin{bmatrix} -l_{\mathrm{max}}\\ -l_{\mathrm{max}} \end{bmatrix} \leq  \begin{bmatrix} \cos\theta_k & \sin\theta_k \\ -\sin\theta_k &  \cos\theta_k\end{bmatrix}
    \begin{bmatrix} p_{x_k} \\ p_{y_k }\end{bmatrix}
    \leq \begin{bmatrix} l_{\mathrm{max}}\\ l_{\mathrm{max}} \end{bmatrix}
\end{align}
$\forall k \in [0, N-1]$, where, $l_{\mathrm{max}}$ is the maximum reachable distance of the swing
foot in both directions on the ground.



\newsec{Maneuverability Constraint.} 
The maneuverability constraint provides a good safety strategy that decelerates the robot's walking speed while turning. It couples the turning rate with longitudinal velocity, as shown below:
\begin{align}
\label{eq:hd_vel_cons}
    \begin{bmatrix} \cos\theta_k & \sin\theta_k \end{bmatrix}
    \begin{bmatrix} v_{x_{k}} \\ v_{y_{k} }\end{bmatrix}
    \leq v_{x_\mathrm{max}} - \frac{\alpha}{\pi}|\omega_k|,
\end{align}
where $\alpha$ is a positive coefficient that balances the turning rate and walking velocity. In our work, we empirically set $\alpha = 3.6$ for the Digit robot. 

\section{Obstacle Avoidance using LDCBF}
\label{sec:dcbf}

We enforce obstacle avoidance by incorporating Discrete Control Barrier Functions (DCBF). We showed in our previous work that regular DCBFs produce nonlinear constraints~\cite{peng2024unified}, which are not amenable for real-time planning. In this section, we propose a novel strategy to express the obstacle avoidance constraint as linear DCBFs (LDCBF).

For the discrete states of the robot $\mathbf{x}_k \in \mathcal{X}\subseteq \mathbb{R}^n$, and discrete control inputs $\mathbf{u}_k \in \mathcal{U} \subseteq \mathbb{R}^m$, if there is a continuous and differentiable function $h: \mathbb{R}^n \rightarrow \mathbb{R}$, the safety set $\mathcal{C}$ and the safety boundary $\partial\mathcal{C}$ of the system may be defined as:
\begin{equation}
\label{eq:safety-set}
\begin{aligned}
    \mathcal{C} &= \{\mathbf{x}_k\in \mathcal{X} |h(\mathbf{x}_k) \geq 0\},\\
    \partial\mathcal{C} &= \{\mathbf{x}_k\in \mathcal{X} |h(\mathbf{x}_k) = 0\}.
\end{aligned}
\end{equation}
Then, $h(\cdot)$ is a discrete control barrier function (DCBF) if there exists a class $\kappa$ function satisfying $0 < \gamma(h(\mathbf{x})) \leq h(\mathbf{x})$, and following conditions can be hold~\cite{agrawal2017discrete,zeng2021safety}:
\begin{align}
\label{eq:cbf_off}
    \forall\text{ } \mathbf{x}_k\in\mathcal{C}.\quad \exists \text{ } \mathbf{u}_k \text{ s.t. } \bigtriangleup h(\mathbf{x}_k, \mathbf{u}_k) \geq -\gamma(h(\mathbf{x}_k)),
\end{align}
where $\bigtriangleup h(\mathbf{x}_k, \mathbf{u}_k) \coloneqq h(\mathbf{x}_{k+1}) - h(\mathbf{x}_k)$. In the discrete domain, $\gamma$ can be also a scalar that $0 < \gamma \leq 1$. So, a DCBF constraint can be written as:
\begin{equation}
\label{eq:cbf_simp}
\begin{aligned}
    \exists \text{ } \mathbf{u}_k &\text{ s.t. } h(\mathbf{x}_{k+1}) + (\gamma - 1)h(\mathbf{x}_k) \geq 0.
\end{aligned}
\end{equation}

\subsection{LDCBF for Path Planning}
\begin{figure}
\centering
\vspace{2mm}
    \includegraphics[trim={0cm 0cm 0cm 0cm},clip,width=1\columnwidth]{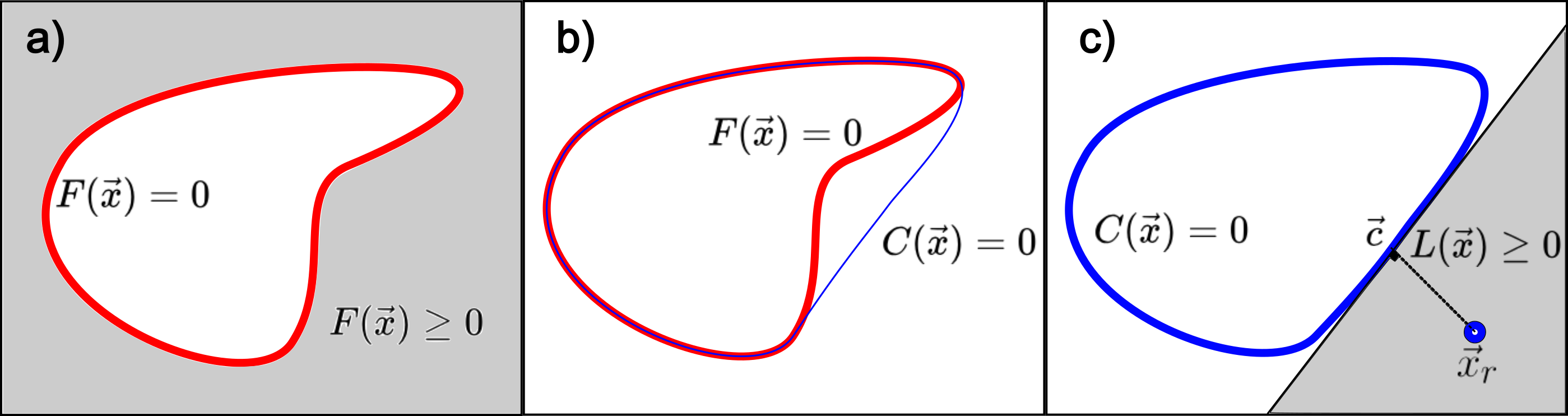}
\caption{a) A function $F(\vec{x})$ represents the outline of an obstacle. The safe path corresponds to the condition $F(\vec{x}) \geq 0$, where $\vec{x}$ represents a position in the map. b) The first step to linearize the DCBF condition is to generate the convex function $C(\vec{x})$ that contains $F(\vec{x})$. c) We use this convex function to consider the robot position $\vec{x}_r$ and find the half-plane formed by the closest point $\vec{c}$ and its normal vector, representing the linear approximation of the safe region.}
\label{fig:F_CBF_Linear}
\end{figure}

To avoid obstacles, we need to compare the position of the robot against the location of the obstacles. For this purpose, we use the vector $\vec{x} := [p_x, p_y]$ that represent a position on the map. Moreover, the selector matrix $S_x$ maps the 3D-LIP states to the robot's position via $\vec{x} = S_x \mathbf{x}$, with
\begin{align}
    S_x = \begin{bmatrix} 1 & 0 & 0 & 0 & 0 \\ 
0 & 0 & 1 & 0 & 0 \end{bmatrix}
\end{align}
In general, for path planning, each obstacle can be represented by a function $F(\vec{x})=0$ as shown in Fig. \ref{fig:F_CBF_Linear}.  Typical representations of this function involve circles and ellipses due to their simpler mathematical form. This function can be constructed such that $F(\vec{x}) < 0$ whenever $\vec{x}$ is inside the obstacle, and $F(\vec{x}) \geq 0$ otherwise. This fact shows that is convenient to choose $h(\mathbf{x}) = F(\vec{x})$ as a DCBF. Even if an obstacle have a complex shape, it is possible to construct a single $h(\mathbf{x})$ as a nonlinear composition of other simpler DCBF~\cite{molnar2023composing, breeden2023compositions}. However, as long as the DCBF function is nonlinear, the DCBF constraint (\ref{eq:cbf_simp}) will also be generally non-linear, which is not amenable for real-time computation.

Given a robot position $\vec{x}_r(t)$ at an instant $t$, it is possible to approximate the safe region with a LDCBF constraint with a half-space provided by the hyperplane $h(\mathbf{x})$ by finding the point $\vec{c}$ that represents the closest point in $F(\vec{x})=0$ to $\vec{x}_r(t)$. The half-space approximation (Fig. \ref{fig:F_CBF_Linear}, c) is given by: 
\begin{align}
\label{eq:hyperplane}
    h(\mathbf{x})= \frac{\nabla  F(\vec{c})^T}{||\nabla F(\vec{c})||} (\vec{x} - \vec{c}) \geq 0
\end{align}
where, $\frac{\nabla  F(\vec{c})^T}{||\nabla F(\vec{c})||}$ represents the normal vector of $F(\vec{x})$ at $\vec{x}=\vec{c}$ that points outwards the obstacle.
\begin{proposition}
If $F(\vec{x})$ is convex, or we use a surrogate convex function $C(\vec{x})$ that contains $F(\vec{x})$, as shown in Fig. \ref{fig:F_CBF_Linear} b), we guarantee that the safe region given by (\ref{eq:hyperplane}) does not contain any unsafe points corresponding to the obstacle.  
\end{proposition}

\begin{proof}
    Assume a differentiable function $F(\vec{x}): \mathbb{R}^2 \rightarrow \mathbb{R}$. If $F(\vec{x})$ is convex, i.e, for any two points $\vec{x}$ and $\vec{y}$ in the obstacle contour, $F(\vec{y}) \geq F(\vec{x}) + \nabla F(\vec{x})^T (\vec{y} - \vec{x})$, then since $F(\vec{x})=F(\vec{y})=0$ we get that $0 \geq \nabla F(\vec{x})^T(\vec{y}-\vec{x})$. Consequently, for any point $\vec{c}$, we observe that the half-space given by $h(\mathbf{x}) = \frac{\nabla F(\vec{c})^T}{ || \nabla F(\vec{c}) ||}(\vec{x}-\vec{c}) \geq 0$ does not cross the obstacle, thus is a safe region. \qedsymbol{}
\end{proof}

\begin{figure}
\centering
    \includegraphics[trim={0cm 0cm 0cm 0cm},clip,width=1\columnwidth]{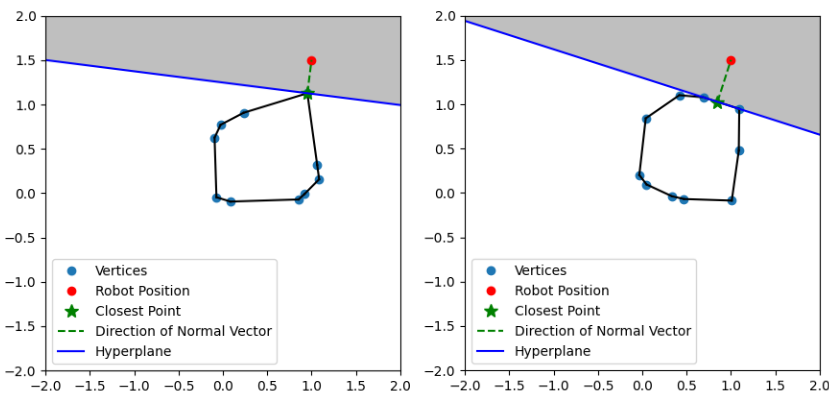}
\caption{Left: The closest point  $\vec{c}$ to the robot is a vertex of the convex hull. Right: The closest point $\vec{c}$ lies within an edge of the convex hull. In both cases, the unit normal vector is calculated as the normalization of the line connecting the closest point to the robot position.}
\label{fig:Liner_DCBF_Cases}
\vspace{-3mm}
\end{figure}

\begin{figure}
\centering
\vspace{2mm}
    \includegraphics[trim={0cm 0cm 0cm 0cm},clip,width=1\columnwidth]{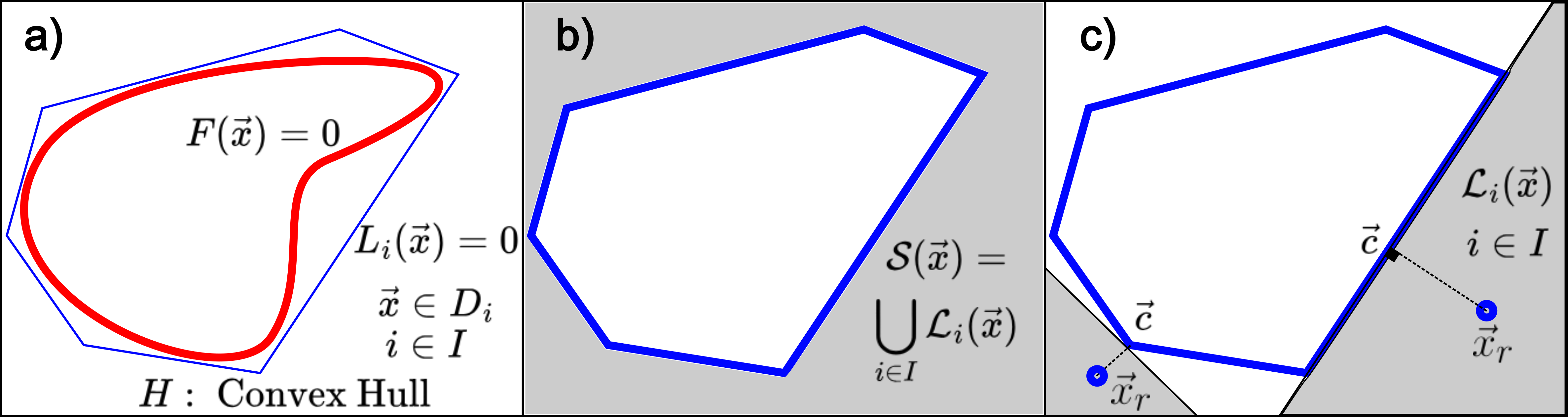}
\caption{a) A general nonlinear $F(\vec{x})=0$ obstacle contour can be approximated by a convex polygon with lines $L_i(\vec{x}) = 0$ delimiting a convex hull. b) Each line produces a safe half-space $\mathcal{L}_i = \{L_i(\vec{x}) \geq 0 \}$. The union of these half-spaces $\mathcal{S}(\vec{x})$ produces the safe region for the robot. c) We simplify $\mathcal{S}(\vec{x})$ considering a unique hyperplane per obstacle depending on the position of the robot $\vec{x}_r$. We compute the closest point  to the robot position $\vec{c}$ and construct the LDCBF as done in (\ref{eq:half-space-convex}).}
\label{fig:F_CBF_Convex}
\end{figure}

Moreover, finding a continuous function $F(\vec{x})$ that represents an obstacle might not be practical. Instead, we can quickly generate a convex polygon that contains the obstacle. In the discontinuous case, the convexity also guarantees that the resulting half-space does not contain any part of the obstacle itself. 
The construction of the hyperplane can be visualized in Fig. \ref{fig:Liner_DCBF_Cases}. The procedure consists of finding the closest point $\vec{c}$ that lives in $F(\vec{x})=0$ and it is closest to the robot position $\vec{x}_r(t)$ at time $t$. This closest point $\vec{c}$ can be either on an edge or be one of the vertices of the convex polygon.  The computation of the normal vector of an edge is $\eta = \frac{\nabla  F(\vec{c})^T}{|| \nabla F(\vec{c})||}$, however, for a vertex, we use $\eta = \frac{\vec{x}_r(t) - \vec{c}}{||\vec{x}_r(t) - \vec{c}||}$, as illustrated in Fig. \ref{fig:F_CBF_Convex}. In either case, the half-space is represented by:
\begin{align}
    \label{eq:half-space-convex}
    h(\mathbf{x}) = \eta^T(\vec{x} - c) \geq 0.
\end{align}

\vspace{-2mm}
\begin{proposition}
Given a linear discrete model that represents the motion of the robot from $x_k$ to $x_{k+1}$ as in (\ref{eq:system_dynamics}) and a DCBF constraint of the form (\ref{eq:cbf_simp}). It can be shown that $h(x_k) = \eta^T (\vec{x}_k - \vec{c})$ can yield a LDCBF constraint if the closest point $\vec{c}$ and the normal vector $\eta$ are approximated as constants for the next time-instant $(k+1)$.      
\end{proposition}

\begin{proof}
We show this by expanding the LDCBF constraint:
\begin{align*}
    h(\mathbf{x}_{k+1}) + (\gamma - 1) h(\mathbf{x}_k) \geq 0 \\
    \eta^T S_x B_L \mathbf{u}_k + \eta^TS(A_L \mathbf{x}_k + (\gamma-1) \mathbf{x}_k) - \eta^T \gamma \vec{c} \geq 0,
\end{align*}
where $S_x B_L \neq 0$ and the constraint is linear in $\mathbf{u}_k$, it can be implemented in a QP-based optimization in real-time.

\end{proof}

\begin{figure}
\centering
    \includegraphics[trim={0cm 0cm 0cm 0cm},clip,width=1\columnwidth]{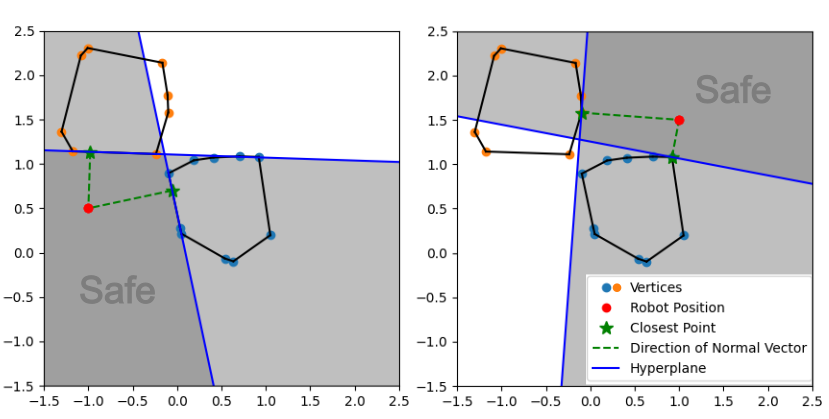}
\caption{Left: A robot starting in the lower left side of the map will experience a safe region composed by the union of half-spaces produced by each obstacle. Right: As the robot moves, the safe region will be updated potentially allowing the robot to reach a target position.}
\label{fig:Linear_DCBF_2Obstacles}
\vspace{-3mm}
\end{figure}

In the case of multiple obstacles, we add one linear constraint per obstacle, producing a safe region given by the intersection of each respective half-space as in Fig. \ref{fig:Linear_DCBF_2Obstacles}. 

\section{Simulation Results}
\label{sec:results}

\begin{figure*}
\centering
\vspace{2mm}
    \includegraphics[trim={0cm 0cm 0cm 0cm},clip,width=1\textwidth]{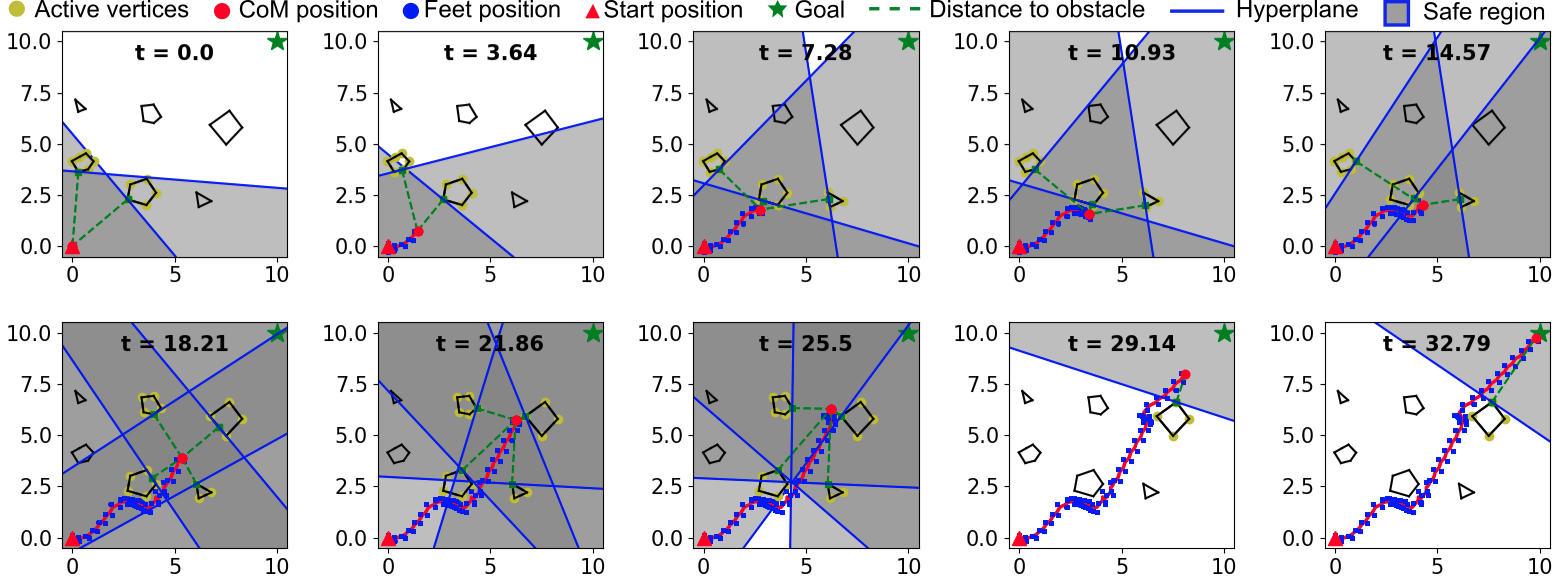}
\caption{Evolution of the active safe half-spaces during the robot motion using the safe MPC framework. The robot starts at (0,0) m and moves toward the goal at (10,10) m. At any given time, only obstacles within 4 meters of the robot are considered active and shown in yellow. The intersection of each corresponding half-space provides the safe region.}
\label{fig:Snapshots}
\vspace{-5mm}
\end{figure*}

This section presents simulation results that demonstrate the effectiveness and performance of the proposed LIP-MPC in navigating clustered environments.
\footnote{A video showing all simulation results can be found in the \url{https://youtu.be/TgTBriqT0Lo}.}.

\subsection{Simulation Setup}

To ensure the safe navigation of the robot, we implemented a hierarchical structure where the LIP-MPC is responsible for generating the next stepping position at 20 Hz update frequency, and a low-level task space controller keeps the CoM height constant, the torso upright and places the swing foot at its desired location at 1 kHz update frequency. We utilized the Agility Robotics' Digit humanoid as our testing platform in simulation. To assess the efficacy and performance of our proposed method, we randomly generated several test environments within the MuJoCo simulator, each featuring eight polygon-shaped obstacles of varying sizes and shapes. In these test scenarios, the starting position was set at $[0,0]$ m, with the goal at $[10,10]$ m. The robot's Center of Mass (CoM) height was maintained at $H=1$ m, the step duration was set to $T=0.4$ s, and the MPC prediction horizon was defined as $N=3$. \tabref{table:2} presents the selected values of weights and limits used throughout all the tests in this paper.

\begin{table}[h]
\vspace{4mm}
\centering
\begin{tabular}{l l  } \hline
Parameters & Value \\ \hline
$[v_{x_\mathrm{min}}, v_{x_\mathrm{max}}]$ & $[-0.1, 0.8] $ m/s \\
$[v_{y_\mathrm{min}}, v_{y_\mathrm{max}}]$ & $[0.1, 0.4]$ m/s\\
$l_{\mathrm{max}}$ & $0.1\sqrt{3}$m \\
$\alpha$ & $1.44$\\
$\gamma$ & $0.3$ \\
\hline
\end{tabular}
\caption{The value of each control parameter used in the simulation throughout this work.}
\label{table:2}
\vspace{-2mm}
\end{table}

\begin{figure}
\centering
    \includegraphics[trim={5.5cm 0cm 4cm 0cm},clip,width=1\columnwidth]{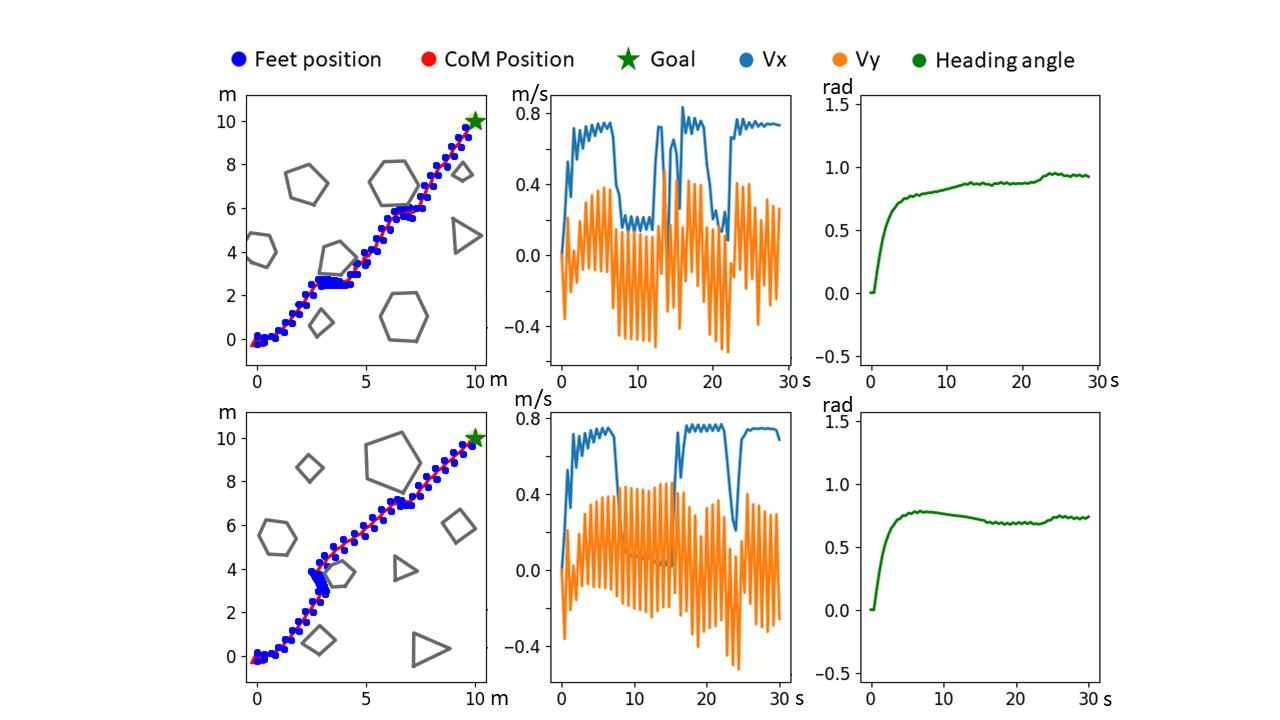}
\caption{The results of the global goal-oriented method in two different cases. The First column shows the robot foot displacements and CoM trajectories. The second column shows the longitudinal and lateral velocity [$m/s$] change over time [s] during the navigation. The third column shows the heading angle changes [$rad$] over time [s].}
\label{fig:LIP-MPC_QP}
\vspace{-1mm}
\end{figure}

\subsection{Global Goal Oriented Planning}
Our approach begins by determining the required turning rate for each prediction step based on the global goal position and then calculating the corresponding foot placement through our linearized LIP-MPC. We tested this method in two randomly generated environments. \figref{fig:Snapshots} illustrates the active LDCBF and dynamically changing safety region from the starting position to the goal in one test. The form and number of LDCBF constraints adapt based on the robot's position. To minimize redundancy and overlap in constraint effects, only obstacles within a 4-meter radius are considered as active LDCBF constraints in our method.

The simulation results for both environments are shown in \figref{fig:LIP-MPC_QP}. The linearized LIP-MPC successfully generated stable stepping positions, guiding the robot safely to the goal without falling. 
Since the target heading angle is computed so that the robot always faces the goal during walking, the heading angle remains relatively constant during the test, as shown in \figref{fig:LIP-MPC_QP}. This strategy reduces the robot's flexibility. When the obstacles are located in the robot's walking direction, this approach avoids obstacles in slow sidewalking rather than with more agile forward walking movements. While bipedal robots are omnidirectional, they move faster and more stably in longitudinal than lateral directions. In these two tests, the robot takes an average of 75 steps with nearly 31 seconds to reach the goal.

\subsection{Sub-goals Oriented Planning}

\begin{figure}
\centering
    \includegraphics[trim={5.5cm 0cm 4cm 0cm},clip,width=1\columnwidth]{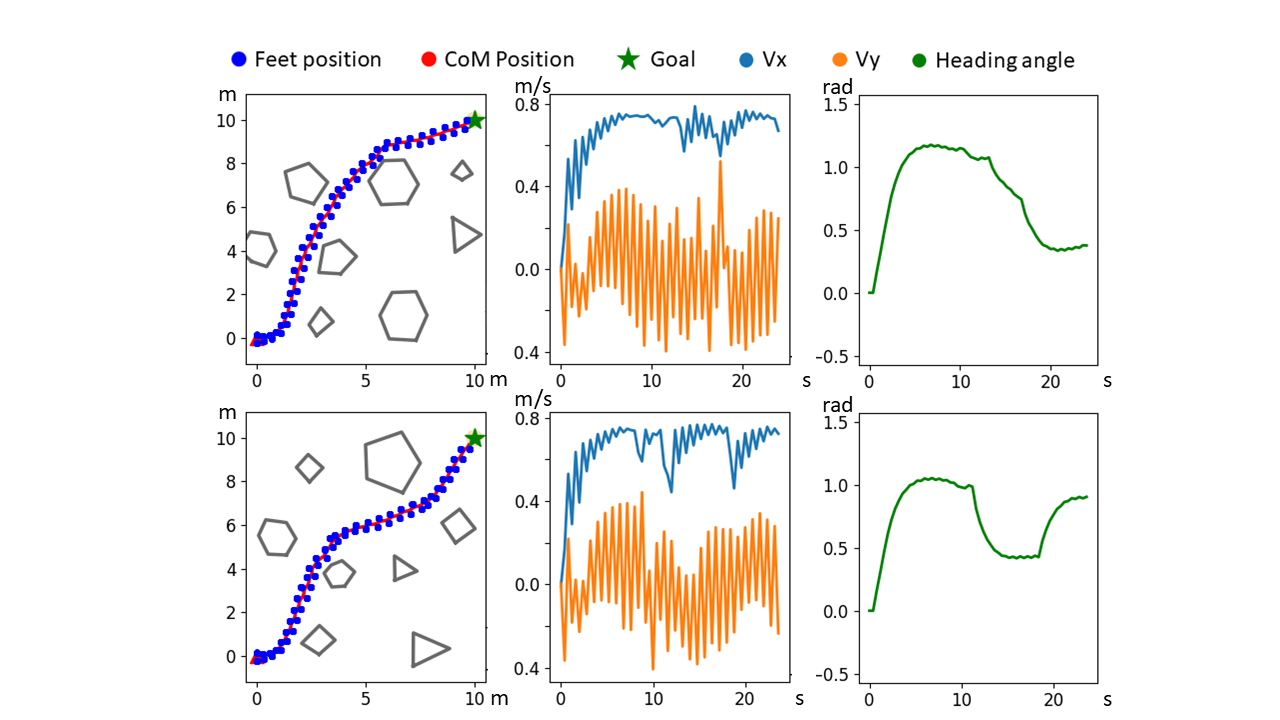}
\caption{The results of the subgoals-oriented method in two different cases. It shows a smoother path (the first column) of the robot, and the longitudinal velocity can keep a high velocity (the second column) compared to the global goal-oriented approach. The third column illustrates the more frequent changes in heading angle.}
\label{fig:LIP-MPC_QP_rrt}
\vspace{-2mm}
\end{figure}

We enhance the navigation framework by introducing sub-goals between the starting position and the final goal. This allows for more flexible steering during the navigation to the goal while preserving the linear form of the LDCBFs. In particular, we employ the Rapidly-exploring Random Tree (RRT) algorithm as a global planner to generate sub-goals that guide the linearized LIP-MPC. The resulting robot CoM trajectories in the same two environments are shown in \figref{fig:LIP-MPC_QP_rrt}. This allows the robot to walk more often in the longitudinal direction, reaching the goal faster with 62 steps in 26 seconds. 
Compared to the previous method, the sub-goal-oriented approach produces a smoother trajectory with a higher average forward velocity and requires fewer steps. 


\section{Conclusion} 
\label{sec:conclusion}

This paper presents a linearized LIP-MPC structure for bipedal locomotion planning. The proposed method begins by determining the turning rate of each step and then obtains a foot placement sequence through a QP-based MPC. Moreover, we also introduce a novel LDCBF with a linear structure applicable to convex obstacles. The results demonstrate the reliability of our method for navigation in various obstacle environments and the feasibility of realizing real-time gait control of bipedal robots. We also discuss the benefits of using a global planner to generate sub-goals such that the robot can achieve more flexible navigation. Despite its effectiveness, there is room for optimizing the pre-computation of turning rates, to ensure safe and optimal steering in clustered environments. Our future work will focus on developing novel model-based and learning-based approaches for optimal steering and hardware realization of the proposed approaches in real-world experiments.

\bibliographystyle{IEEEtran}
\bibliography{references}

\end{document}